\pdfoutput=1

\documentclass[11pt]{article}

\usepackage{acl}
\usepackage{comment}
\usepackage{natbib} 

\usepackage{times}
\usepackage{latexsym}

\usepackage[T1]{fontenc}

\usepackage[utf8]{inputenc}

\usepackage{microtype}

\usepackage{inconsolata}

\usepackage{graphicx}

\title{BIG5-TPoT: Predicting BIG Five Personality Traits, Facets, and Items Through Targeted Preselection of Texts}

\author{
  \textbf{Triet M. Le\textsuperscript{1}},
  \textbf{Arjun Chandra\textsuperscript{2}},
  \textbf{C. Anton Rytting\textsuperscript{1}},
  \textbf{Valerie P. Karuzis\textsuperscript{1}},
\\
  \textbf{Vladimir Rife\textsuperscript{1}},
  \textbf{William A. Simpson\textsuperscript{1}}
\\
\\
  \textsuperscript{1}The University of Maryland Applied Research Laboratory for Intelligence and Security (ARLIS),\\
  \textsuperscript{2}The University of Maryland (UMD), College Park
\\
  \small{
    \textbf{Correspondence:} \href{mailto:trietle@umd.edu}{trietle@umd.edu}
  }
}

\newcommand{\argmax}{{\mathrm{argmax}}}

\begin{document}
\maketitle
\begin{abstract}
Predicting an individual's personalities from their generated texts is a challenging task, especially when the text volume is large. In this paper, we introduce a straightforward yet effective novel strategy called targeted preselection of texts (TPoT). This method semantically filters the texts as input to a deep learning model, specifically designed to predict a Big Five personality trait, facet, or item, referred to as the BIG5-TPoT model. By selecting texts that are semantically relevant to a particular trait, facet, or item, this strategy not only addresses the issue of input text limits in large language models but also improves the Mean Absolute Error and accuracy metrics in predictions for the Stream of Consciousness Essays dataset.
\end{abstract}

\section{Introduction}

Learning to predict one's BIG Five personality traits \cite{goldberg1993structure, mccrae1987validation}  or the Big Five Inventory-2 (BFI2) facets \cite{soto2017next} from text is a form of text classification or regression. Within this framework, textual representation plays a crucial role, as it transforms raw text into a structured form that machine learning algorithms can process. Effective textual representation methods capture the semantic and syntactic nuances of the text, allowing models to understand context and meaning more accurately. This enhances the ability of models to distinguish between different classes, or predict continuous values (in the regression case), based on textual input. Well-represented text ensures that important features are retained and noise is minimized, leading to improved model performance. 

There is a great body of existing work on the representation of texts in natural language processing (NLP), including Latent Semantic Analysis \cite{deerwester_indexing_1990}, Support Vector Machine \cite{hearst_support_1998}, Latent Dirichlet Allocation \cite{blei_latent_2003}, word embedding such as Word2Vec \cite{mikolov_efficient_2013} and GloVE \cite{pennington_glove_2014}, and document embedding (Doc2Vec) \cite{le_distributed_2014}, among others. In this work, we leverage recent advancements in pretrained large language models (LLM) to represent texts for downstream tasks, such as classification and regression. These advancements follow the transformer architectural design in machine translation proposed in \cite{vaswani_attention_2017}. The Bidirectional Encoder Representations from Transformers (BERT) \cite{devlin_bert_2018} is a pretrained LLM that follows this architectural design. Recent advances in generative or multimodal LLMs include ChatGPT-4 from OpenAI, MM1 from Apple, LlaMA-3 from Meta, among others.

For the purpose of showing the power of TPoT, we consider a class of regression models that use pretrained LLMs such as BERT and their finetuned versions to represent the texts. These pretrained LLMs tend to have an input limit of 512 text tokens. We exclude the class of regression models that use a generative LLM such as ChatGPT-4 from OpenAI or LLaMA-3 from Meta, although the same TPoT approach can also be applied to provide inputs to these generative LLMs which have an input token limit of 8,192. 

A substantial amount of research has been conducted on predicting personality traits and facets from texts, and this research evolves with advances in NLP. For a review of the literature, we refer interested readers to \citet*{agastya2019systematic}, \citet*{ahmad2020systematic}, \citet*{zhao2022deep}, and \citet*{singh2024ai}. Here, we focus on research using pretrained LLMs such as BERT to represent texts and follow the work of \citet*{devlin_bert_2018}, \citet*{sun2019fine}, \citet*{leonardi_multilingual_2020}, and \citet*{jain2022personality} for text classification and regression.

The outline of the paper is as follows. In Section \ref{data_sec}, we describe the stream of consciousness Essays dataset used for this study and the process of acquiring ground truth scores for the BIG5 traits and facets. Section \ref{model_sec} rev the standard approach for text regression using a BERT-type pretrained LLM  \cite{leonardi_multilingual_2020}, and the proposed regression models using TPoT for input texts.  Section \ref{result_sec} compares the performance metrics over a ten-fold cross validation of the standard and proposed models against the baseline of always predicting the mean scores. In Section \ref{conclude_sec}, we provide some insights and possible future improvements.

\section{Dataset}\label{data_sec}

In this study, we consider the dataset consisting of 5,810 stream of consciousness essays written by undergraduates at an American university that Dr. James Pennebaker and colleagues collected in 2015-2023, here referred to as the Pennebaker 2015-2023 Essays; a portion of this dataset was used in \citet*{vine2020natural}. To safeguard the identities of these essays’ authors, we used the Microsoft Presidio toolkit to anonymize any Personally Identifiable Information (PII) that they might have included in the essays that they wrote. That is to say, if any personal names, locations, datetimes, or other PII were detected by Presidio’s analyzer tool, that text would be replaced with <PERSON>, <LOCATION>, <DATE\_TIME>, and other such tokens by the Presidio anonymizer. To obtain actual scores for personality traits and facets, students take a personality survey which includes 60 BFI-2 items \cite{soto2017next}, which contain some reversed items. In this survey\footnote{https://www.colby.edu/wp-content/uploads/2013/08/bfi2-form.pdf}, each item is scored based on the author's response: 1 for "disagree strongly," 2 for "disagree a little," 3 for "neutral; no opinion", 4 for "agree a little," and 5 for "agree strongly". The score for a reversed item is the responded score subtracted from 6. Finally, the score for each trait or facet is then the average of the scores of its corresponding items or reversed items.

Below are examples of the 4 BFI-2 item sentences that correspond to the Compassion facet, which falls under the Agreeableness trait. The average of the scores of these 4 items gives the score for the Compassion score. Similarly, the average of the scores of the facets Compassion, Respectfulness, and Trust gives the score for the Agreeableness trait. The same applies to other traits and facets.

\begin{itemize}
\item Item 2: "I am compassionate and have a soft heart." Reverse: "I am unsympathetic and have a hardened heart."
\item Item 17: "I feel little sympathy for others." Reverse: "I feel lots of sympathy for others."
\item Item 32: "I am helpful and unselfish with others." Reverse: "I am callous and selfish with others."
\item Item 47: "I can be cold and uncaring." Reverse: "I am warm and caring."
\end{itemize}

To learn the token statistics of the texts, we use the tokenizer from the pretrained LLM model {\em bert-base-uncased} \cite{devlin_bert_2018} to transform the text from each author to WordPiece tokens \cite{wu2016google}. Table \ref{essay_stat_tab} shows the statistics for the dataset of 5,810 essays on the number of tokens per essay, number of tokens per sentence and the number of sentences per essay. To extract sentences from essays, we use Spacy sentence tokenizer with the language model {\em en\_core\_web\_sm}. To have a complete assessment of the BFI2 traits and facets, an individual has to respond to 60 items in the survey. Thus, we should not expect to get a reliable prediction of all traits and facets from an author's text containing only a single sentence or a few sentences, because they do not capture all the nuances conveyed in the 60 items. Based on the statistics for the number of sentences per essay, we postulate that the Essays dataset is not sufficient as a benchmark, because 75\% of the essays have less than 55 sentences and those sentences may not be relevant to some of the 60 survey items. However, we will show that the TPoT approach can still be applied here to give a better performance.

\begin{table}
    \centering
    \begin{tabular}{lcccc}
        \hline\\
        \multicolumn{3}{r}{\textbf{Percentile}}\\
        \hline\\
         \textbf{Statistics type} & \textbf{25\%} & \textbf{50\%} & \textbf{75\%} & \textbf{95\%}  \\
        \hline\\
        tokens/essay & 650 & 890 & 1131 & 1495\\ 
        tokens/sentence & 12 & 18 & 27 & 44\\
        sentences/essay & 27 & 40 & 55 & 85 \\
        \hline\\
    \end{tabular}
    \caption{Statistics on the number of WordPiece tokens per essay (row 1), number of WordPiece tokens per sentence (row 2) and number of sentences per essay (row 3).}
    \label{essay_stat_tab}
\end{table}

\section{Modeling Approach}\label{model_sec}

\begin{figure}[t]
\includegraphics[width=\columnwidth]{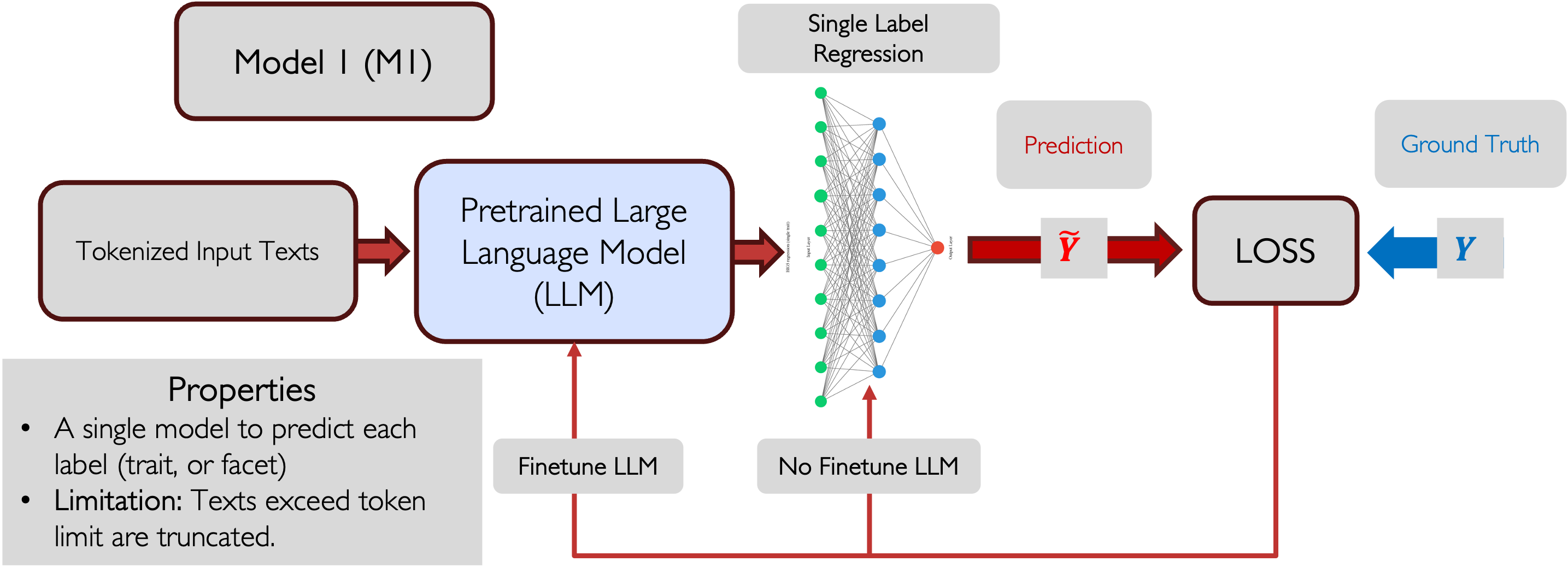}
\caption{Trait, facet, and item regression by finetuning a pretrained BERT or by feature-based (no-finetuning).}\label{m1_fig}
\end{figure}

A common approach to personality regression, using a pretrained LLM to represent the input texts, can be visualized in Figure \ref{m1_fig} \cite{leonardi_multilingual_2020}. In Model 1 (M1), the tokenized text from each author is passed to a pretrained LLM encoder to produce a vector representation. The limit on the number of input tokens $K$ and the dimension $M$ of the vector depend on the LLM being used. The limitation on the number of tokens means that texts with more than $K$ tokens will be truncated. In this study, we take the mean-pooling of the the last hidden state of the encoder to compute an $M$-dimensional vector (document embedding) representing the (at most $K$) input tokens from each author. For a typical LLM \cite{devlin_bert_2018}, $K=512$ and $M=768$. The document embedding is then passed to a 2-layer network  to predict the score $\widetilde{Y}$ for a single trait or facet.

\begin{figure}[t]
\includegraphics[width=\columnwidth]{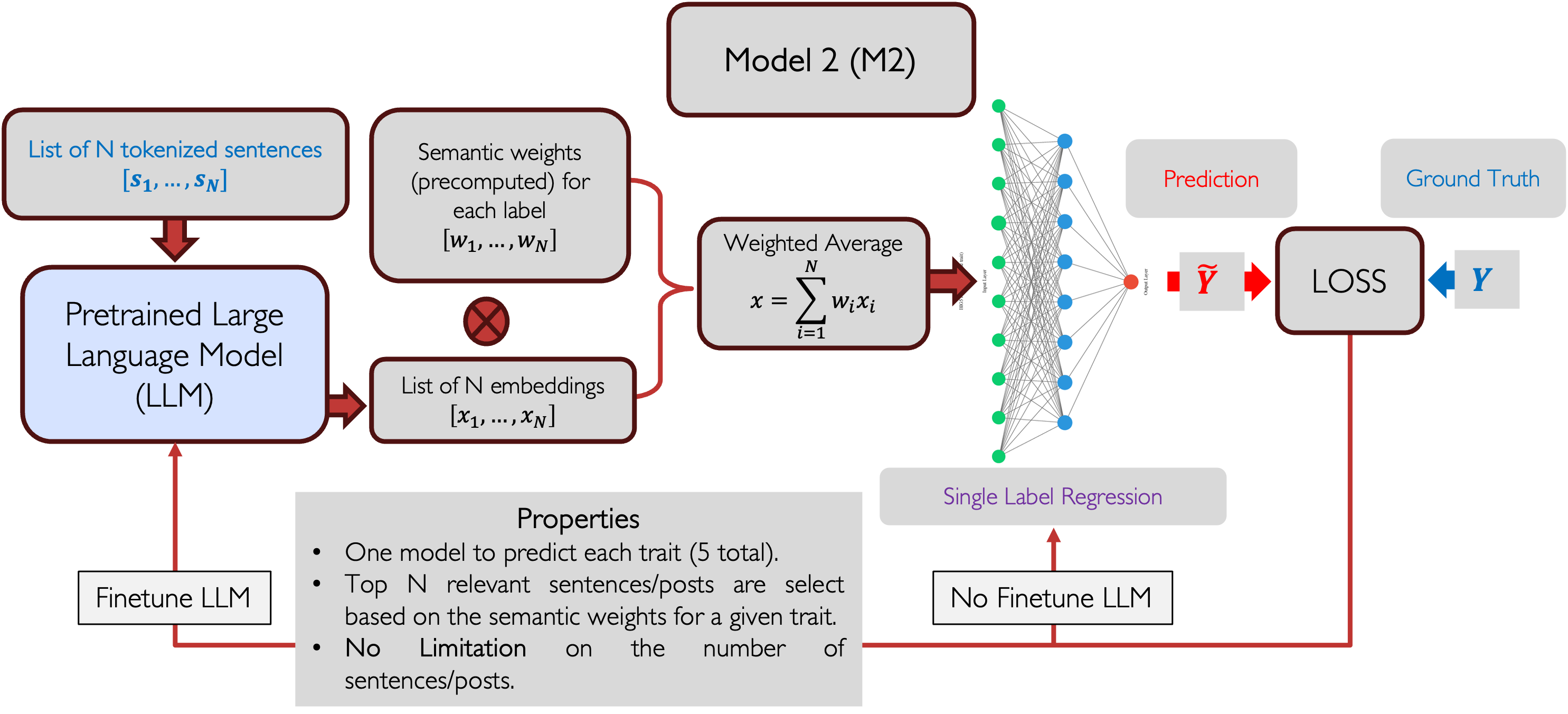}
\caption{Trait, facet, and item regression with TPoT by finetuning a pretrained BERT or by feature-based (no-finetuning).}\label{m2_fig}
\end{figure}

Figure \ref{m2_fig} shows the regression Model 2 (M2) using the proposed TPoT strategy. Note that here, each tokenized sentence is an input to the pretrained LLM, where we don't expect a sentence to have more that 512 tokens. Suppose an author's text has $N$ sentences. As before, for each sentence $s_n$, we use mean-pooling of the last hidden state of the encoder to compute an $M$-dimensional vector $x_n$ (sentence embedding) representing $s_n$. Now, for a given trait or facet, denote by $\{z_1, \cdots, z_J\}$ and $\{z^r_1, \cdots, z^r_J\}$ the item sentence embeddings for the corresponding item sentences and their reverses, respectively. The semantic similarity score $\alpha_n$ \cite{reimers_sentence-bert_2019} of the sentence $s_n$ toward that trait or facet is computed as maximum of the cosine-similarity between $x_n$ and the item sentence embeddings. In other words, let
\begin{eqnarray*} 
\beta_n =\max_{j=1,\cdots, J}\left(\cos(x_n, z_j), \cos(x_n, z^r_j)\right),
\end{eqnarray*}
and $\alpha_n = \max(0,\beta_n)$, which has a minimum value of $0$ and a maximum value of $1$. Let $w_n = \alpha_n/\left(\sum_{i=1}^N \alpha_i\right)$. Note if $\sum_{i=1}^N \alpha_i=0$, then we set $\alpha_n=0$ for all $n$. Then a document embedding $x$ representing $N$ sentences from an author is computed as the weighted average of $x_n$, that is
$$
x = \sum_{n=1}^N w_n x_n.
$$
$x$ is then passed to a 2-layer fully-connected network to predict the score $\widetilde{Y}$ approximating the ground truth score $Y$ for the given trait or facet. Additionally, one can introduce a threshold $\delta$ for $\alpha_n$ to discard sentence $s_n$ when $\alpha_n < \delta$. In the experiments below, we use $\delta=0.2$.

In our experiment, the 2-layer network in Model 1 and Model 2, adopted from \cite{leonardi_multilingual_2020}, consists of a fully connected layer transforming a 768 dimensional vector to a vector of 300 dimensions followed by a non-linear ReLU function and a second fully connected layer transforming a 300 dimensional vector to a single real number representing the predicted score for a trait or facet. We use Huber loss in both models.

In addition to predicting traits or facets, we also consider Model 3 (M3), which predicts each of the 60 BFI2 items using ordinal regression following \cite{cao_rank_2020}. In M3, we apply the same TPoT and the semantic weighted average strategies to compute the document embedding $x$ for each item. In other words, for a fixed item $j$ with the item sentence embedding $z_j$, the semantic score $\alpha_n$ for the sentence embedding $x_n$ is given by
$$
\alpha_n = \max\left(0, \cos(x_n, z_j), \cos(x_n, z^r_j) \right).
$$

 See Appendix \ref{app_a} for a more in depth discussion of ordinal regression. Similar to M1 and M2, we also consider the architecture for 2-layer network to predict the mean of the one dimensional Logistic distribution function.

\section{Experimental Results}\label{result_sec}

For the experiments, we train and test the models on 10 different folds. For each fold, we randomly split the dataset so that 80\% is for training and 20\% is for testing. During training, 10\% of the training set is used for validation. For model performance comparison, we also consider a baseline model which always predicts the mean values of the scores from the training set. When we train a model to predict facets, facet scores are then used to produce the predicted scores for traits. Similarly, when we train a model to predict the scores for each individual item, the item scores are then combined to produce the predicted scores for facets and traits. For all the experiments, we use the no finetune versions of Model 1 and 2 with the LLM {\em sentence-transformers/paraphrase-multilingual-mpnet-base-v2} \cite{reimers_sentence-bert_2019}, which has the maximum number of tokens equal to 512 and the mean pooling of the last hidden state to compute sentence or document embeddings.

Here, we consider two main metrics: 1) Mean absolute error (MAE), and 2) accuracy within a threshold of 0.5. In particular, if the difference between the prediction and the ground truth score is less than or equal to 0.5, then we say the prediction is correct. Otherwise it is incorrect. The accuracy metric is the sum of all the correct predictions in the test set divided by the size of the test set. Additionally, in all of the experiments below, we use the threshold of $\delta=0.2$ for semantic similarity scores.

For simplicity, we use the following acronyms for the 5 traits and the 15 facets (see Table \ref{acro_tab}).

\begin{table}
    \centering
    \resizebox{\linewidth}{!}{%
    \begin{tabular}{lccc}
    \hline\\
         \textbf{Acronym} & \textbf{Trait} & \textbf{Facet} \\
        \hline\\
        \textbf{O}  & Open-Mindedness & -\\
        \textbf{O\_Int} & - & Intellectual Curiosity \\
        \textbf{O\_Eas} & - & Aesthetic Sensitivity \\
        \textbf{O\_Cre} & - & Creative Imagination \\
        \hline\\
        \textbf{C} & Conscientiousness & - \\
        \textbf{C\_Org} & - & Organization \\
        \textbf{C\_Pro} & - & Productiveness \\
        \textbf{C\_Res} & - & Responsibility \\
        \hline\\
        \textbf{E} & Extraversion   & - \\
        \textbf{E\_Soc} & - & Sociability \\
        \textbf{E\_Ass} & - & Assertiveness \\
        \textbf{E\_Ene} & - & Energy Level \\
        \hline\\
        \textbf{A} & Agreeableness & - \\
        \textbf{A\_Com} & - & Compassion \\
        \textbf{A\_Res} & - & Respectfulness \\
        \textbf{A\_Tru} & - & Trust \\
        \hline \\
        \textbf{N} & Negative Emotionality & - \\
        \textbf{N\_Anx} & - & Anxiety \\
        \textbf{N\_Dep} & - & Depression \\
        \textbf{N\_Emo} & - & Emotional Volatility \\
        \hline\\
    \end{tabular}}
    \caption{Acronyms for traits and facets.}
    \label{acro_tab} 
\end{table}

In the first experiment, we apply Model 1 and Model 2 to predict each of the five traits on the test sets from the ten folds. In this case, Model 1 is proposed in \cite{leonardi_multilingual_2020} to predict traits from social media posts. Table \ref{trait_met_tab}-\ref{trait_acc_tab} show the mean and standard deviation for the MAE and accuracy metrics within the threshold of $\epsilon=0.5$ for each trait and each model. We observe that the proposed Model 2 outperforms Model 1 and the baseline in both metrics. 

\begin{table}
    \centering
    \resizebox{\linewidth}{!}{%
    \begin{tabular}{lccc}
        \hline\\
        \multicolumn{4}{c}{\textbf{Trait Prediction: MAE $\downarrow$}}\\
        \hline\\
         & \textbf{Baseline} & \textbf{Model 1} & \textbf{Model 2 (BIG5-TPoT)} \\
        \hline\\
        \textbf{O}  & $0.520 \pm 0.009$ & $0.515\pm 0.010$& ${\bf 0.503}\pm 0.011$ \\
        \textbf{C} & $0.550 \pm 0.008$  & $0.548 \pm 0.010$  & ${\bf 0.538} \pm 0.009$ \\
        \textbf{E} & $0.587 \pm 0.011$  & $0.578 \pm 0.011$  & ${\bf 0.564} \pm 0.012$ \\
        \textbf{A} & $0.490 \pm 0.011$  & $0.492 \pm 0.014$  & ${\bf 0.480} \pm 0.010$ \\
        \textbf{N} & $0.652 \pm 0.016$  & $0.628 \pm 0.014$  & ${\bf 0.608} \pm 0.016$ \\
        \hline\\
    \end{tabular}}
    \caption{MAE metrics for the test sets from the 10 folds showing the mean and standard deviation for each trait and model. $\downarrow$ indicates smaller value is better.}
    \label{trait_met_tab}
\end{table}

\begin{table}
    \centering
    \resizebox{\linewidth}{!}{%
    \begin{tabular}{lccc}
        \hline\\
        \multicolumn{4}{c}{\textbf{Trait Prediction: ACC $(\epsilon=0.5) \uparrow$}}\\
        \hline\\
         & \textbf{Baseline} & \textbf{Model 1} & \textbf{Model 2 (BIG5-TPoT)} \\
        \hline\\
        \textbf{O} & $0.532 \pm 0.015$ & $0.550 \pm 0.009$ & ${\bf 0.560} \pm 0.015$ \\
        \textbf{C} & $0.523 \pm 0.008$ & $0.526 \pm 0.010$ & ${\bf 0.529} \pm 0.008$ \\
        \textbf{E} & $0.499 \pm 0.014$ & $0.502 \pm 0.009$ & ${\bf 0.522} \pm 0.011$ \\
        \textbf{A} & $0.567 \pm 0.017$ & $0.567 \pm 0.013$ & ${\bf 0.589} \pm 0.013$ \\
        \textbf{N} & $0.452 \pm 0.014$ & $0.471 \pm 0.017$ & ${\bf 0.481} \pm 0.019$ \\
        \hline\\
    \end{tabular}}
    \caption{Accuracy metrics  within a threshold of $\epsilon=0.5$ for the test sets from the 10 folds showing the mean and standard deviation for each trait and model. $\uparrow$ indicates larger value is better.}
    \label{trait_acc_tab}
\end{table}

In the second experiment, we apply Model 1 and Model 2 to predict each of the 15 facets on the test sets from the ten folds. The 5 trait scores are computed by averaging the corresponding facets. Tables \ref{facet_met_tab}-\ref{facet_acc_tab} show the mean and standard deviation for the MAE and accuracy metrics within the threshold of $\epsilon=0.5$ for each trait and each model. We observe again that the proposed Model 2 outperforms Model 1 and the baseline on both metrics, except two instances (C\_Res and A\_Res) where the baseline model performs better than Model 2 on MAE.

\begin{table}
    \centering
    \resizebox{\linewidth}{!}{%
    \begin{tabular}{lccc}
        \hline\\
        \multicolumn{4}{c}{\textbf{Facet Prediction: MAE $\downarrow$}}\\
        \hline\\
         & \textbf{Baseline} & \textbf{Model 1} & \textbf{Model 2 (BIG5-TPoT)} \\
        \hline\\
        \textbf{O} & $0.520 \pm 0.009$ & $0.512 \pm 0.009$ & ${\bf 0.502} \pm 0.010$ \\
        \textbf{O\_Int} & $0.572 \pm 0.010$ & $0.571 \pm 0.010$ & ${\bf 0.562} \pm 0.011$ \\
        \textbf{O\_Eas} & $0.725 \pm 0.011$ & $0.704 \pm 0.008$ & ${\bf 0.690} \pm 0.012$ \\
        \textbf{O\_Cre} & $0.645 \pm 0.014$ & $0.648 \pm 0.011$ & ${\bf 0.641} \pm 0.014$ \\
        \hline\\
        \textbf{C} & $0.550 \pm 0.008$ & $0.545 \pm 0.009$ & ${\bf 0.536} \pm 0.008$ \\
        \textbf{C\_Org} & $0.770 \pm 0.013$ & $0.769 \pm 0.015$ & ${\bf 0.755} \pm 0.011$ \\
        \textbf{C\_Pro} & $0.661 \pm 0.014$ & $0.658 \pm 0.019$ & ${\bf 0.645} \pm 0.014$ \\
        \textbf{C\_Res} & ${\bf 0.588} \pm 0.006$ & $0.599 \pm 0.010$ & $0.592 \pm 0.011$ \\
        \hline\\
        \textbf{E} & $0.587 \pm 0.011$ & $0.575 \pm 0.010$ & ${\bf 0.564} \pm 0.010$ \\
        \textbf{E\_Soc} & $0.795 \pm 0.015$ & $0.787 \pm 0.015$ & ${\bf 0.779} \pm 0.015$ \\
        \textbf{E\_Ass} & ${\bf 0.722} \pm 0.016$ & $0.726 \pm 0.017$ & $0.723 \pm 0.015$ \\
        \textbf{E\_Ene} & $0.666 \pm 0.013$ & $0.657 \pm 0.013$ & ${\bf 0.637} \pm 0.013$ \\
        \hline\\
        \textbf{A} & $0.490 \pm 0.011$ & $0.487 \pm 0.011$ & ${\bf 0.477} \pm 0.013$ \\
        \textbf{A\_Com} & $0.615 \pm 0.008$ & $0.616 \pm 0.007$ & ${\bf 0.606} \pm 0.011$ \\
        \textbf{A\_Res} & ${\bf 0.568} \pm 0.014$ & $0.574 \pm 0.013$ & $0.570 \pm 0.015$ \\
        \textbf{A\_Tru} & $0.623 \pm 0.013$ & $0.634 \pm 0.012$ & ${\bf 0.622} \pm 0.014$ \\
        \hline\\
        \textbf{N} & $0.652 \pm 0.016$ & $0.625 \pm 0.016$ & ${\bf 0.606} \pm 0.015$ \\
        \textbf{N\_Anx} & $0.699 \pm 0.015$ & $0.674 \pm 0.017$ & ${\bf 0.659} \pm 0.015$ \\
        \textbf{N\_Dep} & $0.818 \pm 0.013$ & $0.787 \pm 0.015$ & ${\bf 0.763} \pm 0.016$ \\
        \textbf{N\_Emo} & $0.819 \pm 0.017$ & $0.802 \pm 0.016$ & ${\bf 0.793} \pm 0.018$ \\
        \hline\\
    \end{tabular}}
    \caption{MAE metrics showing the mean and standard deviation from 10 folds for the  baseline, Model 1 and the proposed Model 2. $\downarrow$ indicates smaller value is better.}
    \label{facet_met_tab}
\end{table}

\begin{table}
    \centering
    \resizebox{\linewidth}{!}{%
    \begin{tabular}{lccc}
        \hline\\
        \multicolumn{4}{c}{\textbf{Facet Prediction: ACC $(\epsilon=0.5) \uparrow$}}\\
        \hline\\
         & \textbf{Baseline} & \textbf{Model 1} & \textbf{Model 2 (BIG5-TPoT)} \\
        \hline\\
        \textbf{O} & $0.532 \pm 0.015$ & $0.548 \pm 0.013$ & ${\bf 0.561} \pm 0.012$ \\
        \textbf{O\_Int} & $0.474 \pm 0.008$ & $0.495 \pm 0.015$ & ${\bf 0.507} \pm 0.012$ \\
        \textbf{O\_Eas} & $0.410 \pm 0.014$ & $0.425 \pm 0.010$ & ${\bf 0.430} \pm 0.010$ \\
        \textbf{O\_Cre} & $0.440 \pm 0.015$ & $0.448 \pm 0.015$ & ${\bf 0.453} \pm 0.017$ \\
        \hline\\
        \textbf{C} & $0.523 \pm 0.008$ & $0.527 \pm 0.011$ & ${\bf 0.530} \pm 0.009$ \\
        \textbf{C\_Org} & $0.361 \pm 0.011$ & $0.371 \pm 0.013$ & ${\bf 0.381} \pm 0.008$ \\
        \textbf{C\_Pro} & $0.445 \pm 0.013$ & $0.450 \pm 0.013$ & ${\bf 0.462} \pm 0.014$ \\
        \textbf{C\_Res} & ${\bf 0.505} \pm 0.012$ & $0.491 \pm 0.010$ & $0.496 \pm 0.012$ \\
        \hline\\
        \textbf{E} & $0.499 \pm 0.014$ & $0.510 \pm 0.010$ & ${\bf 0.520} \pm 0.012$ \\
        \textbf{E\_Soc} & $0.382 \pm 0.012$ & $0.384 \pm 0.014$ & ${\bf 0.392} \pm 0.018$ \\
        \textbf{E\_Ass} & $0.395 \pm 0.014$ & $0.400 \pm 0.015$ & ${\bf 0.399} \pm 0.009$ \\
        \textbf{E\_Ene} & $0.437 \pm 0.011$ & $0.442 \pm 0.015$ & ${\bf 0.460} \pm 0.012$ \\
        \hline\\
        \textbf{A} & $0.567 \pm 0.017$ & $0.571 \pm 0.010$ & ${\bf 0.594} \pm 0.014$ \\
        \textbf{A\_Com} & $0.447 \pm 0.013$ & $0.460 \pm 0.011$ & ${\bf 0.472} \pm 0.011$ \\
        \textbf{A\_Res} & $0.470 \pm 0.019$ & $0.484 \pm 0.016$ & ${\bf 0.490} \pm 0.018$ \\
        \textbf{A\_Tru} & $0.465 \pm 0.013$ & $0.471 \pm 0.015$ & ${\bf 0.477} \pm 0.017$ \\
        \hline\\
        \textbf{N} & $0.452 \pm 0.014$ & $0.476 \pm 0.019$ & ${\bf 0.480} \pm 0.018$ \\
        \textbf{N\_Anx} & $0.414 \pm 0.014$ & $0.440 \pm 0.019$ & ${\bf 0.451} \pm 0.010$ \\
        \textbf{N\_Dep} & $0.354 \pm 0.008$ & $0.361 \pm 0.011$ & ${\bf 0.372} \pm 0.014$ \\
        \textbf{N\_Emo} & $0.353 \pm 0.014$ & $0.359 \pm 0.014$ & ${\bf 0.362} \pm 0.016$ \\
        \hline\\
    \end{tabular}}
    \caption{MAE metrics showing the mean and standard deviation from 10 folds for the  baseline, Model 1 and the proposed Model 2. $\uparrow$ indicates larger value is better.}
    \label{facet_acc_tab}
\end{table}

In the third experiment, we apply the proposed Model 2 to predict each of the 60 BFI2 item scores using ordinal regression (see Appendix \ref{app_a} for the technical discussion). The item scores are combined to compute the scores for each of the 5 traits and 15 facets. Tables \ref{item_met_tab}-\ref{item_acc_tab} show the performance of the proposed model being trained to predict only traits, facets or items. Both models trained to predict facets and items perform comparatively similar on MAE. However, there is a significant improvement in the accuracy metric for the model trained to predict items.

\begin{table}
    \centering
    \resizebox{\linewidth}{!}{%
    \begin{tabular}{lccc}
        \hline\\
        \multicolumn{4}{c}{\textbf{Model 2 (BIG5-TPoT): MAE $\downarrow$}}\\
        \hline\\
         & \textbf{Trait Prediction} & \textbf{Facet Prediction} & \textbf{Item Prediction}\\
        \hline\\
        \textbf{O} & $0.503\pm 0.011$ & ${\bf 0.502} \pm 0.010$ & $0.509 \pm 0.010$ \\
        \textbf{O\_Int} & - & $0.562 \pm 0.011$ &${\bf 0.560} \pm 0.013$ \\
        \textbf{O\_Eas} & - & ${\bf 0.690} \pm 0.012$ &${\bf 0.690} \pm 0.011$ \\
        \textbf{O\_Cre} & - & ${\bf 0.641} \pm 0.014$ & $0.649 \pm 0.012$ \\
        \hline\\
        \textbf{C} & $0.538 \pm 0.009$ & $0.536 \pm 0.008$ & ${\bf 0.535} \pm 0.009$ \\
        \textbf{C\_Org} & - & $0.755 \pm 0.011$ &${\bf 0.752} \pm 0.011$ \\
        \textbf{C\_Pro} & - & $0.645 \pm 0.014$ & ${\bf 0.642} \pm 0.016$ \\
        \textbf{C\_Res} & - & $0.592 \pm 0.011$ & ${\bf 0.587} \pm 0.010$ \\
        \hline\\
        \textbf{E} & $0.564 \pm 0.012$ & $0.564 \pm 0.010$ & ${\bf 0.563} \pm 0.011$ \\
        \textbf{E\_Soc} & - & $0.779 \pm 0.015$ & ${\bf 0.773} \pm 0.016$ \\
        \textbf{E\_Ass} & - & $0.723 \pm 0.015$ & ${\bf 0.722} \pm 0.014$ \\
        \textbf{E\_Ene} & - & $0.637 \pm 0.013$ & ${\bf 0.634} \pm 0.012$ \\
        \hline\\
        \textbf{A} & $0.480 \pm 0.010$ & ${\bf 0.477} \pm 0.013$ & $0.478 \pm 0.010$ \\
        \textbf{A\_Com} & - & ${\bf 0.606} \pm 0.011$ & $0.607 \pm 0.012$ \\
        \textbf{A\_Res} & - & ${\bf 0.570} \pm 0.015$ & $0.575 \pm 0.015$ \\
        \textbf{A\_Tru} & - & $0.622 \pm 0.014$ & ${\bf 0.616} \pm 0.013$ \\
        \hline\\
        \textbf{N} & $0.608 \pm 0.016$ & ${\bf 0.606} \pm 0.015$ & $0.608 \pm 0.015$ \\
        \textbf{N\_Anx} & - & ${\bf 0.659} \pm 0.015$ & $0.666 \pm 0.018$ \\
        \textbf{N\_Dep} & - & $0.763 \pm 0.016$ & ${\bf 0.758} \pm 0.016$ \\
        \textbf{N\_Emo} & - & ${\bf 0.793} \pm 0.018$ & $0.795 \pm 0.021$ \\
        \hline\\
    \end{tabular}}
    \caption{MAE metrics showing the mean and standard deviation from 10 folds for the proposed Model 2 being trained to predict only traits, facets or items. $\downarrow$ indicates smaller value is better.}
    \label{item_met_tab}
\end{table}

\begin{table}
    \centering
    \resizebox{\linewidth}{!}{%
    \begin{tabular}{lccc}
        \hline\\
        \multicolumn{4}{c}{\textbf{Model 2 (BIG5-TPoT): ACC $(\epsilon=0.5) \uparrow$}}\\
        \hline\\
         & \textbf{Trait Prediction} & \textbf{Facet Prediction} & \textbf{Item Prediction}\\
        \hline\\
        \textbf{O} & $0.560 \pm 0.015$  & $0.561 \pm 0.012$ & ${\bf 0.589} \pm 0.009$ \\
        \textbf{O\_Int} & - & $0.507 \pm 0.012$ & ${\bf 0.608} \pm 0.015$ \\
        \textbf{O\_Eas} & - & $0.430 \pm 0.010$ & ${\bf 0.528} \pm 0.007$ \\
        \textbf{O\_Cre} & - & $0.453 \pm 0.017$ & ${\bf 0.549} \pm 0.012$ \\
        \hline\\
        \textbf{C} & $0.529 \pm 0.008$ & $0.530 \pm 0.009$ & ${\bf 0.568} \pm 0.012$ \\
        \textbf{C\_Org} & - & $0.381 \pm 0.008$ & ${\bf 0.467} \pm 0.011$ \\
        \textbf{C\_Pro} & - & $0.462 \pm 0.014$ & ${\bf 0.557} \pm 0.014$ \\
        \textbf{C\_Res} & - & $0.496 \pm 0.012$ & ${\bf 0.607} \pm 0.014$ \\
        \hline\\
        \textbf{E} & $0.522 \pm 0.011$ & $0.520 \pm 0.012$ & ${\bf 0.555} \pm 0.012$ \\
        \textbf{E\_Soc} & - & $0.392 \pm 0.018$ & ${\bf 0.476} \pm 0.016$ \\
        \textbf{E\_Ass} & - & $0.399 \pm 0.009$ & ${\bf 0.491} \pm 0.015$ \\
        \textbf{E\_Ene} & - & $0.460 \pm 0.012$ & ${\bf 0.555} \pm 0.010$ \\
        \hline\\
        \textbf{A} & $0.589 \pm 0.013$ &$0.594 \pm 0.014$ & ${\bf 0.626} \pm 0.015$ \\
        \textbf{A\_Com} & - & $0.472 \pm 0.011$ & ${\bf 0.576} \pm 0.013$ \\
        \textbf{A\_Res} & - & $0.490 \pm 0.018$ & ${\bf 0.611} \pm 0.014$ \\
        \textbf{A\_Tru} & - & $0.477 \pm 0.017$ & ${\bf 0.574} \pm 0.017$ \\
        \hline\\
        \textbf{N} & $0.481 \pm 0.019$ & $0.480 \pm 0.018$ & ${\bf 0.516} \pm 0.018$ \\
        \textbf{N\_Anx} & - & $0.451 \pm 0.010$ & ${\bf 0.539} \pm 0.015$ \\
        \textbf{N\_Dep} & - & $0.372 \pm 0.014$ & ${\bf 0.461} \pm 0.016$ \\
        \textbf{N\_Emo} & - & $0.362 \pm 0.016$ & ${\bf 0.446} \pm 0.018$ \\
        \hline\\
    \end{tabular}}
    \caption{Accuracy metrics within a threshold of $\epsilon=0.5$ showing the mean and standard deviation from 10 folds for the proposed Model 2 being trained to predict only traits, facets or items. $\uparrow$ indicates larger value is better.}
    \label{item_acc_tab}
\end{table}

\section{Conclusion}\label{conclude_sec}

In this paper, we demonstrate a simple yet effective TPoT strategy, to enhance sentences from an author's text for an improved BIG5 personality prediction, by leveraging the survey items where actual personality scores are acquired. This approach can also be applied to other genres of texts, such as social media posts, to mitigate the limitation on the number of input tokens from LLMs, by analogously treating each post as a {\em sentence}. Additionally, we also show that the performance of the proposed model improves as more information about the author's response to the survey is known. In other words, when the actual item scores are known, the model trained to predict items via ordinal regression performs better than a regression model trained to predict traits or facets. In this paper, we treat trait or facet scores as real numbers in the interval $[1,5]$. But the fact is that they are the average of item scores, which are ordinal. Thus, one can also apply ordinal regression to predict traits and facets, which may provide better accuracy metrics.

\section*{Acknowledgments}
We would like to thank Arjun Chandra, Ashley Lewis, Lauren Liberati, and Pace Ockerman for converting the BFI2 survey items and their reverse to actual sentences. This was part of the 10-week Research for Intelligence \& Security Challenges (RISC) Initiative internship program in the Summer of 2023.

\bibliography{mc_biblio}

\appendix
\section{Ordinal Regression}\label{app_a}
Here, we provide mathematical details of the ordinal regression model for item-level prediction motivated by the CORAL model \cite{cao_rank_2020}. 
\begin{itemize}
\item Fix an item $i$ from 1 to 60.
\item For each author $k$, denote by $x_{k,i}$ the embedding (representation) of the texts from author $k$ with respect to item $i$, and by $y_{k,i}$ the actual ordinal score of item $i$ for author $k$.
\item Let $\theta_0=0.5,\theta_1=1.5,\theta_2=2.5,\theta_3=3.5,\theta_4=4.5$, and $\theta_5=5.5$  be the thresholds. One can update these thresholds during training. For this problem, we keep them fixed.
\item Given $x_{k,i}$, the objective is to learn the conditional probability for the prediction $\widetilde{y}_{k,i}$ for each author $k$ and item $i$: 
$$
P(y_{k,i} < \theta_j|x_{k,i}), \mbox{ for } j = 1, \cdots, 5
$$
with $P(y_{k,i} < \theta_0|x_{k,i}) = 0$. Note that the quantity above is monotone increasing with respect to $j: P(y_{k,i} < \theta_{j-1}|x_{k,i})\le P(y_{k,i} < \theta_{j}|x_{k,i})$. This monotone increasing property provides the ordered consistency for ordinal regression. Once this conditional probability is known, then the conditional probability that $\widetilde{y}_{k,i}$ lies between $\theta_{k-1}$ and $\theta_k$ is given by:
\begin{eqnarray*}
f_{k,i}(j) &:=& P(\theta_{j-1} < \widetilde{y}_{k,i} < \theta_{k} | x_{k,i})\\
&=& P(y_{k,i} < \theta_{j}|x_{k,i}) \\
&-&  P(y_{k,i} < \theta_{j-1}|x_{k,i}).
\end{eqnarray*}
\item The ordinal prediction for $\widetilde{y}_{k,i}$ is given by
$$
\widetilde{y}_{k,i} = \argmax_{j=1,\cdots, 5} f_{k,i}(j).
$$
\end{itemize}{
Following \cite{cao_rank_2020}, we use the cumulative function $\sigma$ of the one dimension logistic distribution to model the condition probability $P(y_{k,i} < \theta_j|x_{k,i})$, which has the form
$$
\sigma(z,\mu_{k,i},s_{k,i})=\frac{1}{1+\exp\left(-(z-\mu_{k,i})/s_{k,i}\right)}
$$
where $\mu_{k,i}$ represents the mean and $s_{k,i}$ determines the scale (width) of $\sigma$.

\begin{figure}[t]
\includegraphics[width=\columnwidth]{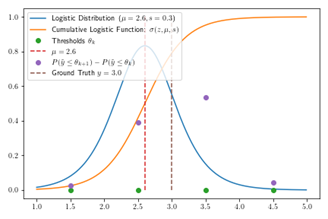}
\caption{Plots of the one-dimensional Logistic distribution and its cumulative function with $\mu = 2.6$ and $s = 0.3$.}\label{log_fig}
\end{figure}

Figure \ref{log_fig} shows the plots of the one-dimensional Logistic distribution and its cumulative function which is determined by the mean $\mu=\mu_{k,i}=2.6)$ and $s=s_{k,i}=0.3$. This example shows that the value $f_{k,i}(j)$ is maximal at $j=3$. This implies the ordinal score $\widetilde{y}_{k,i} = 3$ even though $\mu_{k,i} = 2.6$. Clearly if $\mu_{k,i}$ approaches to $3$ and $s$ approaches to $0$, then $f_{k,i}(3)$ approaches to $1$. One benefit of ordinal regression is the integration of the uncertainty (i.e. the scale variable $s$) in predicting the ordinal score. During training, we update $\mu_{k,i}, s_{k,i}$ and $\widetilde{y}_{k,i}$ by minimizing the sum of two losses. The first loss is between the predicted ordinal scores $\widetilde{y}_{k,i}$ and the ground truth scores $y_{k.i}$ using binary cross entropy loss. The second loss is between the mean $\mu_{k,i}$ and the $y_{k.i}$ using Huber loss.

\end{document}